\documentclass[11pt,a4paper]{article}
\usepackage[hyperref]{udst2018}
\usepackage{times}
\usepackage{amsmath}
\usepackage{amssymb}
\usepackage{latexsym}

\usepackage{booktabs}
\usepackage{xcolor}
\usepackage{rotating}
\usepackage{subfigure}
\usepackage{colortbl}

\usepackage{url}
\usepackage{multirow}

\usepackage{tikz, pgfplots}
\usepackage{pgfplotstable}
\usepackage{paralist}

\definecolor{red}{HTML}{E31A1C}
\definecolor{blue}{HTML}{1F78B4}
\definecolor{green}{HTML}{33A02C}
\definecolor{orange}{HTML}{FF7F00}
\definecolor{purple}{HTML}{6A3D9A}

\definecolor{Gray}{gray}{0.85}
\newcolumntype{z}{>{\columncolor{Gray}}l}

\newcommand{\conllu}{CoNLL-U}
\newcommand{\fone}{$\text{F}_1$}

\aclfinalcopy
\def\aclpaperid{***} %  Enter the Paper ID here for final camera ready copy

\title{Universal Dependency Parsing from Scratch}

\author{
Peng Qi,*
Timothy Dozat,*
Yuhao Zhang,*
Christopher D.\ Manning\\
Stanford University\\
Stanford, CA 94305 \\
  {\tt \{pengqi, tdozat, yuhaozhang, manning\}@stanford.edu}
}

\date{}

\begin{document}
\maketitle

\setlength{\abovedisplayskip}{5pt}
\setlength{\belowdisplayskip}{5pt}
\setlength{\abovedisplayshortskip}{0pt}
\setlength{\belowdisplayshortskip}{0pt}

\renewcommand{\thefootnote}{\fnsymbol{footnote}}
\footnotetext[1]{These authors contributed roughly equally.}
\renewcommand{\thefootnote}{\arabic{footnote}}

\newcommand{\udst}[0]{\emph{CoNLL 2018 UD Shared Task}}

% Outline:
% Tokenizer/segmenter/MWT setup (Peng)
% Lemmatizer setup (Yuhao)
% POS Tagger (Tim)
% - LSTM Sequence labeler, highway connections, ReLU layers
% - For languages with smaller tagsets, affine UPOS classifier, biaffine XPOS/UFeats classifiers
% - For languages with larger tagsets, affine UPOS classifier, shared hidden layer
% - For languages with compositional (array-like) X tagsets, used the feature approach
% Parser (Tim)
% - Same basic setup as DQM17, + highway connections/lemma/UFeats
% - attachment score included P(i<j | edge(j, i)) and P(abs(i-j) | edge(j, i))
% - More dropout

\begin{abstract}

This paper describes Stanford's system at the \udst.
We introduce a complete neural pipeline system that takes raw text as input, and performs all tasks required by the shared task, ranging from tokenization and sentence segmentation, to POS tagging and dependency parsing.
Our single system submission achieved very competitive performance on big treebanks.
Moreover, after fixing an unfortunate bug, our corrected system would have placed the 2\textsuperscript{nd}, 1\textsuperscript{st}, and 3\textsuperscript{rd} on the official evaluation metrics LAS, MLAS, and BLEX, and would have outperformed all submission systems on low-resource treebank categories on all metrics by a large margin.
We further show the effectiveness of different model components through extensive ablation studies.

\end{abstract}

\section{Introduction}

Dependency parsing is an important component in various natural langauge processing (NLP) systems for semantic role labeling \cite{marcheggiani2017encoding}, relation extraction \cite{zhang2018graph}, and machine translation
% \cite{chen2017improved, eriguchi2017learning, wu2017sequence}.
\cite{chen2017improved}.
However, most research has treated dependency parsing in isolation, and largely ignored upstream NLP components that prepare relevant data for the parser, \emph{e.g.}, tokenizers and lemmatizers \cite{udst:overview2017}.
In reality, however, these upstream systems are still far from perfect.

To this end, in our submission to the \udst{}, we built a  raw-text-to-\conllu{} pipeline system that performs all tasks required by the Shared Task \cite{udst:overview}.%
\footnote{We chose to develop a pipeline system mainly because it allows easier parallel development and faster model tuning in a shared task context.}
Harnessing the power of neural systems, this pipeline achieves competitive performance in each of the inter-linked stages: tokenization, sentence and word segmentation, part-of-speech (POS)/morphological features (UFeats) tagging, lemmatization, and finally, dependency parsing.
Our main contributions include:
\begin{compactitem}
    \item New methods for combining symbolic statistical knowledge with flexible, powerful neural systems to improve robustness;
    \item A biaffine classifier for joint POS/UFeats prediction that improves prediction consistency;
    \item A lemmatizer enhanced with an \emph{edit} classifier that improves the robustness of a sequence-to-sequence model on rare sequences; and
    \item Extensions to our parser from \cite{dozat-qi-manning:2017:K17-3} to %incorporate information about
    model linearization.
\end{compactitem}

Our system achieves competitive performance on big treebanks. %, and can be further improved by perfecting individual components.
After fixing an unfortunate bug, the corrected system would have placed the 2\textsuperscript{nd}, 1\textsuperscript{st}, and 3\textsuperscript{rd} on the official evaluation metrics LAS, MLAS, and BLEX,
% in our unofficial evaluation,
and would have outperformed all submission systems on low-resource treebank categories on all metrics by a large margin.
We perform extensive ablation studies to demonstrate the effectiveness of our novel methods, and highlight future directions to improve the system.%
\footnote{To facilitate future research, we make our implementation public at: \url{https://github.com/stanfordnlp/stanfordnlp}.}

% In our previous year's submission to the UD Shared Task, we largely ignored all tasks other than part-of-speech tagging and dependency parsing \cite{dozat-qi-manning:2017:K17-3}.
% In the \udst, however, the two newly introduced metrics---MLAS and BLEX---emphasize upstream processing %focusing
% at the lexical level, such as morphological feature (UFeats) prediction and lemmatization.
% To this end, we built a raw-text-to-\conllu{} pipeline with neural networks that aims to leverage data-driven learning to improve the performance of each of the inter-linked stages: tokenization, sentence and word segmentation, POS/UFeats tagging, lemmatization, and finally, dependency parsing.
% We present new methods for combining symbolic statistical knowledge with flexible, powerful neural systems to improve robustness, as well as improvements to our winning system from last year to further push the boundary of single-system tagging and parsing performance.

\section{System Description}

In this section, we present detailed descriptions for each component of our neural pipeline system, namely the tokenizer, the POS/UFeats tagger, the lemmatizer, and finally the dependency parser.

\subsection{Tokenizer} \label{sec:tok}

To prepare sentences in the form of a list of words for downstream processing, the tokenizer component reads raw text and outputs sentences in the \conllu{} format.
This is achieved with two subsystems: one for joint tokenization and sentence segmentation, and the other for splitting multi-word tokens into syntactic words. %so they can be properly handled by downstream systems.

\paragraph{Tokenization and sentence segmentation.} We treat joint tokenization and sentence segmentation as a unit-level sequence tagging problem.
For most languages, a \emph{unit} of text is a single character; however, in Vietnamese orthography, the most natural units of text are single \emph{syllables}.%
\footnote{In this case, we define a syllable as a consecutive run of alphabetic characters, numbers, or individual symbols, together with any leading white spaces before them.}
We assign one out of five tags to each of these units: end of token (\textsc{eot}), end of sentence (\textsc{eos}), multi-word token (\textsc{mwt}), multi-word end of sentence (\textsc{mws}), and other (\textsc{other}).
We use bidirectional LSTMs (BiLSTMs) as the base model to make unit-level predictions.
At each unit, the model predicts hierarchically: it first decides whether a given unit is at the end of a token with a score $s^{(\text{tok})}$, then classifies token endings into finer-grained categories with two independent binary classifiers: one for sentence ending $s^{(\text{sent})}$, and one for MWT $s^{(\textsc{mwt})}$.

\begin{figure}
  \centering
  \includegraphics[width=.41\textwidth]{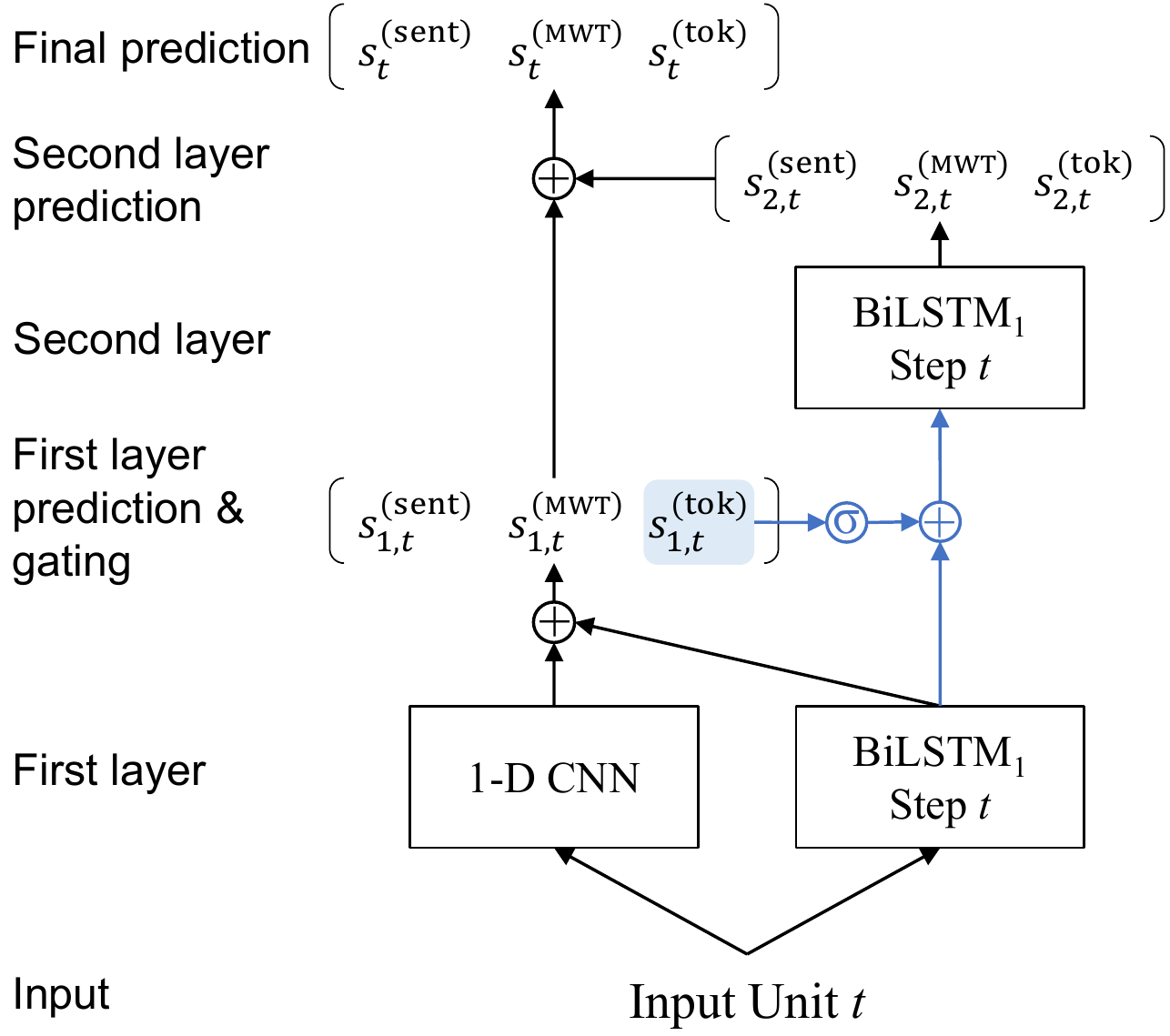}
  \vspace{-1em}
  \caption{Illustration of the tokenizer\slash{}sentence segmenter model. Components in blue represent the gating mechanism between the two layers.} \label{fig:tokenizer}
\end{figure}

Since sentence boundaries and MWTs usually require a larger context to determine (\emph{e.g.}, periods following abbreviations or the ambiguous word ``des'' in French), we incorporate token-level information into a two-layer BiLSTM as follows (see also Figure \ref{fig:tokenizer}). The first layer BiLSTM operates directly on raw units, and makes an initial prediction over the categories.
To help capture local unit patterns more easily, we also combine the first-layer BiLSTM with 1-D convolutional networks, by using a one hidden layer convolutional network (CNN) with ReLU nolinearity at its first layer, giving an effect a little like a residual connection \cite{he2016residual}.
The output of the CNN is simply added to the concatenated hidden states of the BiLSTM for downstream computation:
\begin{align}
\mathbf{h}^{\text{RNN}}_1 = [\overrightarrow{\mathbf{h}_1}, \overleftarrow{\mathbf{h}_1}] &= \text{BiLSTM}_1(\mathbf{x}),\\
\mathbf{h}^{\text{CNN}}_1 &= \text{CNN}(\mathbf{x}),\\
\mathbf{h}_1 &= \mathbf{h}_1^{\text{RNN}} + \mathbf{h}_1^{\text{CNN}}, \\
[\mathbf{s}^{(\text{tok})}_1, \mathbf{s}^{(\text{sent})}_1, \mathbf{s}^{(\textsc{mwt})}_1] &= W_1\mathbf{h}_1,
\end{align}
where $\mathbf{x}$ is the input character representations, and $W_1$ contains the weights and biases for a linear classifier.%
\footnote{We will omit bias terms in affine transforms for clarity.}
For each unit, we concatenate its trainable embedding with a four-dimensional binary feature vector as input, each dimension corresponding to one of the following feature functions: (1) does the unit start with whitespace; (2) does it start with a capitalized letter; (3) is the unit fully capitalized; and (4) is it purely numerical.

To incorporate token-level information at the second layer, we use a gating mechanism to suppress representations at non-token boundaries before propagating hidden states upward:
\begin{align}
\mathbf{g}_1 &=\mathbf{h}_1 \odot \sigma(\mathbf{s}^{(\text{tok})}_1) \\\label{eqn:tokenizer_gating}
\mathbf{h}_2 = [\overrightarrow{\mathbf{h}_2}, \overleftarrow{\mathbf{h}_2}] &= \text{BiLSTM}_2(\mathbf{g}_1), \\
[\mathbf{s}^{(\text{tok})}_2, \mathbf{s}^{(\text{sent})}_2, \mathbf{s}^{(\textsc{mwt})}_2] &= W_2\mathbf{h}_2,
\end{align}
where $\odot$ is an element-wise product broadcast over all dimensions of $\mathbf{h}_1$ for each unit. This can be viewed as a simpler alternative to multi-resolution RNNs \cite{serban2017multiresolution}, where the first-layer BiLSTM operates at the unit level, and the second layer operates at the token level. Unlike multi-resolution RNNs, this formulation is end-to-end differentiable, and can more easily leverage efficient off-the-shelf RNN implementations.

To combine predictions from both layers of the BiLSTM, we simply sum the scores to obtain $\mathbf{s}^{(X)}=\mathbf{s}^{(X)}_1+\mathbf{s}^{(X)}_2$,
%\begin{align}
%\mathbf{s}^{(X)} = \sum_{i=1}^2 \mathbf{s}^{(X)}_i,
%\end{align}
where $X\in \{\text{tok, sent, \textsc{mwt}}\}$.
The final probability over the tags is then
\begin{align}
p_{\textsc{eot}} &= p_{+--} & p_{\textsc{eos}} &= p_{++-}, \\
p_{\textsc{mwt}} &= p_{+-+} & p_{\textsc{mws}} &= p_{+++},
% \sigma(s^{(\text{tok})}-s^{(\text{sent})}-s^{(\textsc{mwt})}),\\
% p_{\textsc{eos}} &= \sigma(s^{(\text{tok})}+s^{(\text{sent})}-s^{(\textsc{mwt})}),\\
% p_{\textsc{mwt}} &= \sigma(s^{(\text{tok})}-s^{(\text{sent})}+s^{(\textsc{mwt})}),\\
% p_{\textsc{mws}} &= \sigma(s^{(\text{tok})}+s^{(\text{sent})}+s^{(\textsc{mwt})}),
\end{align}
where $p_{\pm\pm\pm}=\sigma(\pm s^{\text{(tok)}})\sigma(\pm s^{\text{(sent)}})\sigma(\pm s^{\textsc{(mwt)}})$, and $\sigma(\cdot)$ is the logistic sigmoid function.
$p_{\textsc{other}}$ is simply $\sigma(-s^{\text{(tok)}})$.
The model is trained to minimize the standard cross entropy loss.% to predict the correct tag for each unit.

\paragraph{Multi-word Token Expansion.} The tokenizer\slash{}sentence segmenter produces a collection of sentences, each being a list of tokens, some of which are labeled as multi-word tokens (MWTs).
% \footnote{Our system annotates the output \conllu{} file by appending a special \texttt{MWT=Yes} feature to the last column.}
We must expand these MWTs into the underlying syntactic words they correspond to (\emph{e.g.}, ``im'' to ``in dem'' in German), in order for downstream systems to process them properly.
To achieve this, we take a hybrid approach to combine symbolic statistical knowledge with the power of neural systems.

The symbolic statistical side is a frequency lexicon. Many languages, like German, have only a handful of rules for expanding a few MWTs.
We leverage this information by simply counting the number of times a MWT is expanded into different sequences of words in the training set, and retaining the most frequent expansion in a dictionary to use at test time.
When building this dictionary, we lowercase all words in the expansions to improve robustness.
However, this approach would fail for languages with rich clitics, a large set of unique MWTs, and/or complex rules for MWT expansion, such as Arabic and Hebrew. We capture this by introducing a powerful neural system.

Specifically, we train a sequence-to-sequence model using a BiLSTM encoder with an attention mechanism \cite{bahdanau2014neural} in the form of a multi-layer perceptron (MLP).
Formally, the input multi-word token is represented by a sequence of characters $x_1, \ldots, x_I$, and the output syntactic words are represented similarly as a sequence of characters $y_{1}, \ldots, y_J$, where the words are separated by space characters.
Inputs to the RNNs are encoded by a shared matrix of character embeddings $E$.
Once the encoder hidden states $\mathbf{h}^{\text{enc}}$ are obtained with a single-layer BiLSTM, each decoder step is unrolled as follows:
\begin{align}
\mathbf{h}^{\text{dec}}_j &= \text{LSTM}_{\text{dec}}(E_{y_{j-1}}, \mathbf{h}^{\text{dec}}_{j-1}),\\
\alpha_{ij} &\propto \exp(\mathbf{u}^{\top}_\alpha \tanh (W_\alpha[\mathbf{h}^{\text{dec}}_j, \mathbf{h}^{\text{enc}}_i])),\\
\mathbf{c}_{j} &= \sum_i \alpha_{ij} \mathbf{h}^{\text{enc}}_i, \\
P(y_{j}&=w|y_{<j}) \propto \mathbf{u}_w^\top \tanh (W[\mathbf{h}^{\text{dec}}_j, \mathbf{c}_j]).
\end{align}
Here, $w$ is a character index in the output vocabulary, $y_0$ a special start-of-sequence symbol in the vocabulary, and $\mathbf{h}^{\text{dec}}_0$ the concatenation of the last hidden states of each direction of the encoder.

To bring the symbolic and neural systems together, we train them separately and use the following protocol during evaluation:
for each MWT, we first look it up in the dictionary, and return the expansion recorded there if one can be found.
If this fails, we retry by lowercasing the incoming token.
If that fails again, we resort to the neural system to predict the final expansion.
This allows us to not only account for languages with flexible MWTs patterns (Arabic and Hebrew), but also leverage the training set statistics to cover both languages with simpler MWT rules, and MWTs in the flexible languages seen in the training set without fail.
%The result of this hybrid approach is
This results in a high-performance, robust system for multi-word token expansion.

\subsection{POS/UFeats Tagger}
% TODO maybe describe the character model a bit more?
Our tagger follows closely that of \cite{dozat-qi-manning:2017:K17-3}, with a few extensions. As in that work, the core of the tagger is a highway BiLSTM \cite{srivastava2015highway} with inputs coming from the concatenation of three sources: (1) a pretrained word embedding, from the word2vec embeddings provided with the task when available \citep{mikolov2013distributed}, and from fastText embeddings otherwise \citep{bojanowski2016enriching}; (2) a trainable frequent word embedding, for all words that occurred at least seven times in the training set; and (3) a character-level embedding, generated from a unidirectional LSTM over characters in each word. UPOS is predicted by first transforming each word's BiLSTM state with a fully-connected (FC) layer, then applying an affine classifier:
\begin{align}
  \mathbf{h}_i &= \text{BiLSTM}_i^{(\text{tag})}(\mathbf{x}_1, \ldots, \mathbf{x}_n) ,\\
  \mathbf{v}^{(\text{u})}_i &= \text{FC}^{(\text{u})}(\mathbf{h}_i) ,\\
  %\mathbf{s}^{(\text{u})}_i &= W^{(\text{u})}\mathbf{v}^{(\text{u})}_i ,\\
  %P\big(y^{(\text{u})}_{ik} | X\big) &= \text{softmax}_k\big(\mathbf{s}^{(\text{u})}_i\big) .
  P\big(y^{(\text{u})}_{ik} | X\big) &= \text{softmax}_k\big(W^{(\text{u})}\mathbf{v}^{(\text{u})}_i \big) .
\end{align}
To predict XPOS, we similarly start with transforming the BiLSTM states with an FC layer. In order to further ensure consistency between the different tagsets (\emph{e.g.},\ to avoid a \textsc{verb} UPOS with an \textsc{nn} XPOS), we use a \emph{biaffine} classifier, conditioned on a word's XPOS state as well as an embedding for its gold (at training time) or predicted (at inference time) UPOS tag $y_{i*}^{(\text{u})}$:
\begin{align}
  \mathbf{v}^{(\text{x})}_i &= \text{FC}^{(\text{x})}(\mathbf{h}_i) ,\\
%   \mathbf{e}^{(\text{u})}_i &= E^{(\text{u})}_{y_{i*}^{(\text{u})}} ,\\
%   \mathbf{s}^{(\text{x})}_i &= [\mathbf{e}_i^{(\text{u})}, 1]^\top \mathbf{U}^{(\text{x})}[\mathbf{v}^{(\text{x})}_i, 1] ,\\
  \mathbf{s}^{(\text{x})}_i &= [E^{(\text{u})}_{y_{i*}^{(\text{u})}} , 1]^\top \mathbf{U}^{(\text{x})}[\mathbf{v}^{(\text{x})}_i, 1] ,\\
  P\big(y_{ik}^{(\text{x})} | y_{i*}^{(\text{u})}, X\big) &= \text{softmax}_k\big(\mathbf{s}^{(\text{x})}_i\big) .
\end{align}
UFeats is predicted analogously with separate parameters for each individual UFeat tag. The tagger is also trained to minimize the cross entropy loss.

Some languages have composite XPOS tags, yielding a very large XPOS tag space
(\emph{e.g.}, Arabic and Czech).
For these languages, the biaffine classifier requires a prohibitively large weight tensor $\mathbf{U}^{(\text{x})}$.
For languages that use XPOS tagsets with a fixed number of characters, % each representing a feature of the word,
we classify each character of the XPOS tag in the same way we classify each UFeat.
For the rest, instead of taking the biaffine approach, %to encouraging consistency,
we simply share the FC layer between all three affine classifiers, hoping that the learned features for one will be used by another.

\subsection{Lemmatizer}

For the lemmatizer, we take a very similar approach to that of the multi-word token expansion component introduced in Section \ref{sec:tok} with two key distinctions customized to lemmatization.

First, we build two dictionaries from the training set, one from a (word, UPOS) pair to the lemma, and the other from the word itself to the lemma.
During evaluation, the predicted UPOS is used.
When the UPOS-augmented dictionary fails, we fall back to the word-only dictionary before resorting to the neural system.
In looking up both dictionaries, the word is never lowercased, because case information is more relevant in lemmatization than in MWT expansion.

Second, we enhance the neural system with an \textit{edit} classifier that shortcuts the prediction process to accommodate rare, long words, on which the decoder is more likely to flounder.
The concatenated encoder final states are put through an FC layer with ReLU nonlinearity and fed into a 3-way classifier, which predicts whether the lemma is (1) exactly identical to the word (\emph{e.g.}, URLs and emails), (2) the lowercased version of the word (\emph{e.g.}, capitalized rare words in English that are not proper nouns), or (3) in need of the sequence-to-sequence model to make more complex edits to the character sequence.
During training time, we assign the labels to each word-lemma pair greedily in the order of identical, lowercase, and sequence decoder, and train the classifier jointly with the sequence-to-sequence lemmatizer.
At evaluation time, predictions are made sequentially%
%through each component
, \emph{i.e.}, the classifier first determines whether any shortcut can be taken, before the sequence decoder model is used if needed.

\subsection{Dependency Parser}
The dependency parser also follows that of \cite{dozat-qi-manning:2017:K17-3} with a few augmentations. The highway BiLSTM takes as input pretrained word embeddings, frequent word and lemma embeddings, character-level word embeddings, summed XPOS and UPOS embeddings, and summed UFeats embeddings. %Like the tagger, it includes highway connections.
In \cite{dozat-qi-manning:2017:K17-3}, unlabeled attachments are predicted by scoring each word $i$ and its potential heads with a biaffine transformation
% In our submission last year% \cite{dozat-qi-manning:2017:K17-3}
% , unlabeled attachment between a token at index $i$ and its potential heads was computed by applying one fully connected layer to the token (the \emph{edge-dep} state) and another to the potential heads (the \emph{edge-head} state), then scoring each possible edge with a biaffine function: %This was then optimized with cross-entropy.
\begin{align}
  \mathbf{h}_t &= \text{BiLSTM}_t^{(\text{parse})}(\mathbf{x}_1, \ldots, \mathbf{x}_n) , \label{eqn:dbs-1} \\
  \mathbf{v}^{(\text{ed})}_i, \mathbf{v}^{(\text{eh})}_j &= \text{FC}^{(\text{ed})}(\mathbf{h}_i), \text{FC}^{(\text{eh})}(\mathbf{h}_j) ,\\
  s^{(\text{e})}_{ij} &= [\mathbf{v}_j^{(\text{eh})}, 1]^\top U^{(\text{e})}[\mathbf{v}^{(\text{ed})}_i, 1] ,\\
  &= \text{Deep-Biaff}^{(\text{e})}(\mathbf{h}_i, \mathbf{h}_j) , \label{eqn:dbs-3}\\
  P\big(y_{ij}^{(\text{e})}| X\big) &= \text{softmax}_j\big(\mathbf{s}^{(\text{e})}_i\big) ,\label{eqn:Pyij}
\end{align}
where $\mathbf{v}_i^{\textrm{(ed)}}$ is word $i$'s \emph{edge-dependent} representation and $\mathbf{v}_i^{\textrm{(eh)}}$ its \emph{edge-head} representation.
%That approach to finding the head for a given token doesn't
This approach, however, does not explicitly take into consideration relative locations of heads and dependents during prediction; instead, such predictive location information must be implicitly learned by the BiLSTM. Ideally, we would like the model to explicitly condition on $(i-j)$, namely the dependent $i$ and its potential head $j$'s location relative to each other, in modeling $p(y_{ij})$.%
\footnote{Henceforth we omit the (e) superscript and $X$.}
%Ideally, rather than simply computing the edge probability $P(y_{ij})$, the model should learn to condition this prediction on $(i-j)$, namely the relative location of the words.%

%learn $P(y_{ij} | (i-j))$, the probability of an edge between token $i$ and dependent $j$ given their locations relative to each other.
Here, we motivate one way to build this into the model. First we factorize the relative location of word $i$ and head $j$ into their \emph{linear order} and the \emph{distance} between them, \emph{i.e.}, $P(y_{ij} | \text{sgn}(i - j), \text{abs}(i - j))$, where sgn$(\cdot)$ is the sign function. Applying Bayes' rule and assuming conditional independence, we arrive at the following
\begin{align}
  \label{lin-dist}&P(y_{ij} | \text{sgn}(i-j),\text{abs}(i-j))\propto\\
  &\quad P(y_{ij})P(\text{sgn}(i-j) | y_{ij})P(\text{abs}(i-j) | y_{ij})\nonumber .
\end{align}
In a language where heads always follow their dependents, $P(\text{sgn}(i-j) = 1 | y_{ij})$ would be extremely low, heavily penalizing rightward attachments. Similarly, in a language where dependencies are always short, $P(\text{abs}(i-j) \gg 0 | y_{ij})$ would be extremely low, penalizing longer edges.

$P(y_{ij})$ can remain the same as computed in Eq.~(\ref{eqn:Pyij}).
$P(\text{sgn}(i-j) | y_{ij})$ can be computed similarly  with a deep biaffine scorer (cf.\ Eqs.~(\ref{eqn:dbs-1})--(\ref{eqn:dbs-3})) over the recurrent states.
This results in the score of $j$ preceding $i$; flipping the sign wherever $i$ precedes $j$ turns this into the log odds of the observed linearization.
Applying the sigmoid function then turns it into a probability:
\begin{align}
  s_{ij}^{(\text{l})} &= \text{Deep-Biaff}^{(\text{l})}(\mathbf{h}_i, \mathbf{h}_j) ,\\
  s_{ij}^{\prime(\text{l})} &= \text{sgn}(i-j)s_{ij}^{(\text{l})} ,\\
  P(\text{sgn}(i-j) |& y_{ij}) = \sigma\big(s_{ij}^{\prime(\text{l})}\big) \label{eqn:linearization} .
\end{align}
This can be effortlessly incorporated into the edge score %$s_{ij}^{(\text{e})}$
by adding in the log of this probability $-\log(1+\exp(-s_{ij}^{\prime(\text{l})}))$. Error is not backpropagated to this submodule through the final attachment loss; instead, it is trained with its own cross entropy, with error only computed on gold edges. This ensures that the model learns the conditional probability \emph{given a true edge}, rather than just learning to predict the linear order of two words.

For $P(\text{abs}(i-j) | y_{ij})$, we use another deep biaffine scorer to generate a distance score. Distances are always no less than $1$, so we apply $1+\text{softplus}$ to predict the distance between $i$ and $j$ when there's an edge between them:
\begin{align}
  s_{ij}^{(\text{d})} &= \text{Deep-Biaff}^{(\text{d})}(\mathbf{h}_i, \mathbf{h}_j) ,\\
  s_{ij}^{\prime(\text{d})} &= 1+\text{softplus}\big(s_{ij}^{(\text{d})}\big) .
\end{align}
where $\text{softplus}(x) = \log(1+\exp(x))$.
The distribution of edge lengths in the treebanks roughly follows a Zipfian distribution, %with a long tail,
to which the Cauchy distribution is closely related, only the latter is more stable for values at or near zero. Thus, rather than %directly
modeling the probability of an arc's length, we can use the Cauchy distribution to model the probability of an arc's \emph{error} in predicted length, namely how likely it is for the predicted distance and the true distance to have a difference of $\delta_{ij}^{(\text{d})}$:
\begin{align}
  \text{Zipf}(k; \alpha, \beta) &\propto (k^{\alpha}/{\beta})^{-1} ,\\
  \text{Cauchy}(x; \gamma) &\propto (1+x^2/{\gamma})^{-1}\\
  \delta_{ij}^{(\text{d})} &= \text{abs}(i-j) - s_{ij}^{\prime(\text{d})} ,\\
  P(\text{abs}(i-j) | y_{ij}) &\propto (1+\delta_{ij}^{2(\text{d})} / 2)^{-1} \label{eqn:distance} .
\end{align}
When the difference $\delta_{ij}^{(\text{d})}$ is small or zero, there will be effectively no penalty; but when the model expects a significantly longer or shorter arc than the observed distance between $i$ and $j$, it is discouraged from assigning an edge between them. As with the linear order probability, the log of the distance probability %$-\log (1+\delta_{ij}^{2(\text{d})}/2)$
is added to the edge score% $s^{(\text{e})}_{ij}$
, and trained with its own cross-entropy on gold edges.%
\footnote{Note that the penalty assigned to the edge score in this way is proportional to $\ln \delta_{ij}^{(\text{d})}$ for high $\delta_{ij}^{(\text{d})}$;
using a Gamma or Poisson distribution to model the distance directly, or using a normal distribution instead of Cauchy, respectively, assigns penalties roughly proportional to $\delta_{ij}^{(\text{d})}$, $\ln\Gamma(\delta_{ij}^{(\text{d})})$, and $\delta_{ij}^{2(\text{d})}$.
Thus, the Cauchy is more numerically stable during training.}

At inference time, the Chu-Liu/Edmonds algorithm \citep{chu1965shortest,edmonds1967optimum} is used to ensure a maximum spanning tree. Dependency relations are assigned to gold (at training time) or predicted (at inference time) edges $y^{(\text{e})}_{i*}$ using another deep biaffine classifier, following \cite{dozat-qi-manning:2017:K17-3} with no augmentations:%. Each word is put through one FC layer, and its head word is put through a separate one. The results are then used as input to the biaffine relation classifier:
\begin{align}
  \mathbf{s}_{i}^{(\text{r})} &= \text{Deep-Biaff}^{(\text{r})}\big(\mathbf{h}_i, \mathbf{h}_{y_{i*}^{(\text{e})}}\big) ,\\
  P\big(y_{ik}^{(\text{r})} | y_{i*}^{(\text{e})}\big) &= \text{softmax}_k\big(\mathbf{s}_i^{(\text{r})}\big) .
\end{align}

\section{Training Details}
Except where otherwise stated, our system is a pipeline: given a document of raw text, the tokenizer/sentence segmenter/MWT expander first splits it into sentences of syntactic words; the tagger then assigns UPOS, XPOS and UFeat tags to each word; the lemmatizer takes the predicted word and UPOS tag and outputs a lemma; finally, the parser takes all annotations as input and predicts the head and dependency label for each word.

All components are trained with early stopping on the dev set when applicable.
When a dev set is unavailable, we split the training set into an approximately 7-to-1 split for training and development.
All components (except the dependency parser) are trained and evaluated on the development set assuming all related components had oracle implementations.
This means the tokenizer/sentence segmenter assumes all correctly predicted MWTs will be correctly expanded, the MWT expander assumes gold word segmentation, and all downstream tasks assume gold word segmentation, along with gold annotations of all prerequisite tasks.
The dependency parser is trained with predicted tags and morphological features from the POS/UFeats tagger.

\paragraph{Treebanks without training data.} For treebanks without training data, we adopt a heuristic approach for finding replacements.
Where a larger treebank in the same language is available (\emph{i.e.}, all PUD treebanks and Japanese-Modern), we used the models from the largest treebank available in that language.
Where treebanks in related languages are available (as determined by language families from Wikipedia), we use models from the largest treebank in that related language.
We ended up choosing the models from English-EWT for Naija (an English-based pidgin), Irish-IDT for Breton (both are Celtic), and Norwegian-Nynorsk for Faroese (both are West Scandinavian).
For Thai, since it uses a different script from all other languages, we use UDPipe 1.2 for all components.

\paragraph{Hyperparameters.} The tokenizer\slash sentence segmenter %\footnote{For Latin-ITTB only, we found it beneficial to sentence segmentation to always treat colons and semicolons as sentence breaks, and used this heuristic in evaluation.}
uses BiLSTMs with 64d hidden states in each direction and takes 32d character embeddings as input.
During training, we employ dropout to the input embeddings and hidden states at each layer with $p=.33$.
We also randomly replace the input unit with a special \texttt{<UNK>} unit with $p=.33$, which would be used in place of any unseen input at test time.
We add noise to the gating mechanism in Eq.\ (\ref{eqn:tokenizer_gating}) by randomly setting the gates to 1 with $p=.02$ and setting its temperature to 2 to make the model more robust to tokenization errors at test time.
% \footnote{This is easily implemented as a dropout mechanism $(1-\text{dropout}(\sigma(-\mathbf{s}_1^{(tok)}), 0.02))$}
Optimization is performed with Adam \cite{kingma2015adam} with an initial learning rate of .002 for up to 20,000 steps, and whenever dev performance deteriorates, as is evaluated every 200 steps after the 2,000$^{\text{th}}$ step, the learning rate is multiplied by $.999$.
For the convolutional component we use filter sizes of 1 and 9, and for each filter size we use 64 channels (same as one direction in the BiLSTM).
The convolutional outputs are concatenated in the hidden layer, before an affine transform is applied to serve as a residual connection for the BiLSTM.
For the MWT expander, we use BiLSTMs with 256d hidden states in each direction as the encoder, a 512d LSTM decoder, 64d character embeddings as input, and dropout rate $p=.5$ for the inputs and hidden states.
Models are trained up to 100 epochs with the standard Adam hyperparameters, and the learning rate is annealed  similarly every epoch after the 15$^{\text{th}}$ epoch  by a factor of $0.9$.
Beam search of beam size 8 is employed in evaluation.

The lemmatizer uses BiLSTMs with 100d hidden states in each direction of the encoder, 50d character embeddings as input, and dropout rate $p=.5$ for the inputs and hidden states.
The decoder is an LSTM with 200d hidden states.
%For the decoder it uses an LSTM with 200d hidden states and a maximum decoding length of 50.
During training we jointly minimize (with equal weights) the cross-entropy loss of the edit classifier and the negative log-likelihood loss of the seq2seq lemmatizer.
Models are trained up to 60 epochs with standard Adam hyperparameters.

The tagger and parser share most of their hyperparameters.
We use 75d uncased frequent word and lemma embeddings% for words and lemmas that occur seven or more times in the training set
, and 50d POS tag and UFeat embeddings.
Pretrained embeddings and character-based word representations %LSTM attention states
are both transformed to be 125d.
During training, all embeddings are randomly replaced with a \texttt{<drop>} symbol with $p=.33$.
We use 2-layer 200d BiLSTMs for the tagger and 3-layer 400d BiLSTMs for the parser.
We employ dropout in all feedforward connections with $p=.5$ and all recurrent connections \citep{gal2016dropout} with $p=.25$ (except $p=.5$ in the tagger BiLSTM).
All classifiers use 400d FC layers (except 100d for UFeats) with the ReLU nonlinearity.
We train the systems with Adam ($\alpha=.003$, $\beta_1=.9$, $\beta_2=.95$) until dev accuracy decreases, at which point we switch to AMSGrad \citep{reddi2018convergence} until 3,000 steps pass with no dev accuracy increases.

\section{Results}

\begin{table*}
\scriptsize
\centering
\subfigure[Results on all treebanks]{
\begin{tabular}{lllllllllllzzz}
\toprule
System & Tokens & Sent & Words & Lemmas & UPOS & XPOS & UFeats & AllTags & UAS & CLAS & \textbf{LAS} & \textbf{MLAS} & \textbf{BLEX} \\
\midrule
Stanford & 96.19 & 76.55 & 95.99 & 88.32 & 89.01 & 85.51 & 85.47 & 79.71 & 76.78 & 68.73 & 72.29 & 60.92 & 64.04\\
Reference & 98.42$^\dagger$ & 83.87$^\dagger$ & 98.18$^\ddagger$ & 91.24$^\star$ & 90.91$^\ddagger$ & 86.67$^*$ & 87.59$^\ddagger$ & 80.30$^*$ & 80.51$^\dagger$ & 72.36$^\dagger$ & 75.84$^\dagger$ & 61.25$^*$ & 66.09$^\star$\\
\midrule
$\Delta$ & --2.23 & --7.32 & --2.19 & --2.92 & --1.90 & --1.16 & --2.12 & --0.59 & --3.73 & --3.63 & --3.55 & --0.33 & --2.05\\
\midrule
Stanford+ & 97.42 & 85.46 & 97.23 & 89.17 & 89.95 & 86.50 & 86.20 & 80.36 & 79.04 & 70.39 & 74.16 & 62.08 & 65.28\\
\midrule
$\Delta$ & --1.00 & \textbf{+1.59} & --0.95 & --2.07 & --0.96 & --0.17 & --1.39 & \textbf{+0.06} & --1.47 & --1.97 & --1.68 & \textbf{+0.83} & --0.81\\
\bottomrule
\end{tabular}}
\subfigure[Results on big treebanks only]{
\begin{tabular}{lllllllllllzzz}
\toprule
System & Tokens & Sent & Words & Lemmas & UPOS & XPOS & UFeats & AllTags & UAS & CLAS & \textbf{LAS} & \textbf{MLAS} & \textbf{BLEX} \\
\midrule
Stanford & 99.43 & 89.52 & 99.21 & 95.25 & 95.93 & 94.95 & 94.14 & 91.50 & 86.56 & 79.60 & 83.03 & 72.67 & 75.46\\
Reference & 99.51$^\dagger$ & 87.73$^\dagger$ & 99.16$^\dagger$ & 96.08$^\star$ & 96.23$^\dagger$ & 95.16$^\dagger$ & 94.11$^*$ & 91.45$^*$ & 87.61$^\dagger$ & 81.29$^\dagger$ & 84.37$^\dagger$ & 71.71$^*$ & 75.83$^\star$\\
\midrule
$\Delta$ & --0.08 & \textbf{+1.79} & \textbf{+0.05} & --0.83 & --0.30 & --0.21 & \textbf{+0.03} & \textbf{+0.05} & --1.05 & --1.69 & --1.34 & \textbf{+0.96} & --0.37\\
%\midrule
%Stanford+ & 99.43 & 89.54 & 99.23 & 95.27 & 95.94 & 94.96 & 94.15 & 91.51 & 86.57 & 79.60 & 83.04 & 72.67 & 75.45 \\
%\midrule
%$\Delta$ &  \\
\bottomrule
\end{tabular}}
\caption{Evaluation results (\fone{}) on the test set, on all treebanks and big treebanks only. For each set of results on all metrics, we compare it against results from reference systems. A reference system is the top performing system on that metric if we are not top, or the second-best performing system on that metric. Reference systems are identified by superscripts ($\dagger$: HIT-SCIR, $\ddagger$: Uppsala, $\star$:~TurkuNLP, $*$:~UDPipe Future). Shaded columns in the table indicate the three official evaluation metrics. ``Stanford+'' is our system after a bugfix evaluated unofficially; for more details please see the main text.} \label{tab:results}
\end{table*}

The main results are shown in Table \ref{tab:results}.
As can be seen from the table, our system achieves competitive performance on nearly all of the metrics when macro-averaged over all treebanks.
Moreover, it achieves the top performance on several metrics when evaluated only on big treebanks, showing that our systems can effectively leverage statistical patterns in the data.
Where it is not the top performing system, our system also achieved competitive results on each of the metrics on these treebanks.
This is encouraging considering that our system is comprised of single-system components, whereas some of the best performing teams used ensembles (\emph{e.g.}, HIT-SCIR \cite{che18towards}).

When taking a closer look, we find that our UFeats classifier is very accurate on these treebanks as well.
Not only did it achieve the top performance on UFeats \fone, but also it helped the parser achieve top MLAS as well on big treebanks, even when the parser is not the best-performing as evaluated by other metrics.
We also note the contribution from our consistency modeling in the POS tagger/UFeats classifier: in both settings the individual metrics (UPOS, XPOS, and UFeats) achieve a lower advantage margin over the reference systems when compared to the AllTags metric, showing that these reference systems, though sometimes more accurate on each individual task, are not as consistent as our system overall.

The biggest disparity between the all-treebanks and big-treebanks results comes from sentence segmentation.
After inspecting the results on smaller treebanks and double-checking our implementation, we noticed issues with how we processed data in the tokenizer that negatively impacted generalization on these treebanks.%
\footnote{Specifically, our tokenizer was originally designed to be aware of newlines (\texttt{\textbackslash{}n}) in double newline-separated paragraphs, but we accidentally prepared training and dev sets for low resource treebanks by putting each sentence on its own line in the text file.
This resulted in the sentence segmenter overfitting to relying on newlines. In later experiments, we replaced all in-paragraph whitespaces with space characters.}
This is devastating for these treebanks, as all downstream components process words at the sentence level.
%We hope to fix this issue, retrain our tokenizer component, and report new results on all treebanks once test data is available.

\begin{table}
\centering
\small
\begin{tabular}{llzzz}
\toprule
Treebanks & System & \textbf{LAS} & \textbf{MLAS} & \textbf{BLEX} \\
\midrule
\multirow{2}{*}{Small} & Stanford+ & \textbf{83.90} & \textbf{72.75} & \textbf{77.30} \\
& Reference & 69.53$^\dagger$ & 49.24$^\ddagger$ & 54.89$^\ddagger$ \\
\midrule
\multirow{2}{*}{Low-Res} & Stanford+ & \textbf{63.20} & \textbf{51.64} & \textbf{53.58} \\
& Reference & 27.89$^\star$ & 6.13$^\star$ & 13.98$^\star$ \\
\midrule
\multirow{2}{*}{PUD} & Stanford+ & \textbf{82.25} & \textbf{74.20} & \textbf{74.37}\\
& Reference & 74.20$^\dagger$ & 58.75$^*$ & 63.25$^\bullet$ \\
\bottomrule
\end{tabular}
\caption{Evaluation results (\fone{}) on low-resource treebank test sets. Reference systems are identified by symbol superscripts ($\dagger$: HIT-SCIR, $\ddagger$: ICS PAS, $\star$: CUNI x-ling, $*$: Stanford, $\bullet$: TurkuNLP).} \label{tab:lowres}
\end{table}

We fixed this issue, and trained new tokenizers with all hyperparameters identical to our system at submission.
We further built an unofficial evaluation pipeline, which we verified achieves the same evaluation results as the official system, and evaluated our entire pipeline by \emph{only} replacing the tokenizer.
As is shown in Table \ref{tab:results}, the resulting system (Stanford+) is much more accurate overall, and we would have ranked 2\textsuperscript{nd}, 1\textsuperscript{st}, and 3\textsuperscript{rd} on the official evaluation metrics LAS, MLAS, and BLEX, respectively.%
\footnote{We note that the only system that is more accurate than ours on LAS is HIT's ensemble system, and we achieve very close performance to their system on MLAS (only 0.05\% \fone{} lower, which is likely within the statistical variation reported in the official evaluation).}
On big treebanks, all metrics changed within only 0.02\% \fone{} and are thus not included.
On small treebanks, however, this effect is more pronounced: as is shown in Table \ref{tab:lowres}, our corrected system outperforms all submission systems on all official evaluation metrics on all low-resource treebanks by a large margin.

\section{Analysis}

In this section, we perform ablation studies on the new approaches we proposed for each component, and the contribution of each component to the final pipeline.
For each component, we assume access to an oracle for all other components in the analysis, and show their efficacy on the dev sets.%
\footnote{We perform treebank-level paired bootstrap tests for each ablated system against the top performing system in ablation with $10^5$ bootstrap samples, and indicate statistical significance in tables with symbol superscripts (*:$p < 0.05$, **:$p<0.01$, ***:$p < 0.001$).}
For the ablations on the pipeline, we report macro-averaged \fone{} on the test set.

\begin{table}
\centering
\small
\begin{tabular}{lccc}
\toprule
System & Tokens & Sentences & Words \\
\midrule
Stanford+ & 99.46$^{\phantom{*}}$ & 91.33$^{\phantom{***}}$ & \textbf{99.27}$^{\phantom{*}}$ \\
$-$ \textit{gating} & \textbf{99.47}$^{\phantom{*}}$ & \textbf{91.34}$^{\phantom{***}}$ & \textbf{99.27}$^{\phantom{*}}$\\
$-$ \textit{conv} & 99.45$^{\phantom{*}}$ & 91.03$^{\phantom{***}}$ & 98.67$^{\phantom{*}}$\\
$-$ \textit{seq2seq} & --$^{\phantom{*}}$ & --$^{\phantom{***}}$ & 98.97$^{\phantom{*}}$\\
$-$ \textit{dropout} & 99.22$^{*}$ & 88.78$^{***}$ & 98.98$^{*}$\\
\bottomrule
\end{tabular}
\caption{Ablation results for the tokenizer. All metrics in the table are macro-averaged dev \fone{}.} \label{tab:tok_ablation}
\end{table}

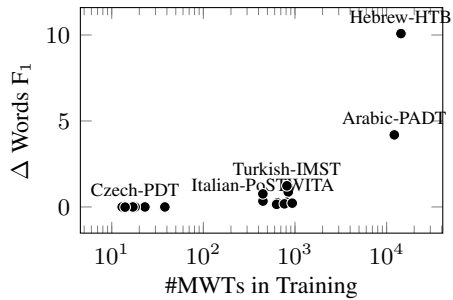
\begin{figure}
    \centering
    \setlength{\abovecaptionskip}{-4pt}
    \begin{tikzpicture}[font=\small]
  % \begin{axis}[nodes near coords,enlargelimits=0.2]
  %   \addplot+[only marks,
  %   point meta=explicit symbolic]
  %   coordinates {
  %   (0.5,0.2) [(1)]
  %   (0.2,0.1) [(2)]
  %   (0.7,0.6) [(3)]
  %   (0.35,0.4) [(4)]
  %   (0.65,0.1) [(5)]
  %   };
  %   \end{axis}
  \begin{semilogxaxis}[width=2.5in, height=1.8in, enlargelimits=0.15,
    x label style={at={(axis description cs:0.5,0.03)},anchor=north},
    y label style={at={(axis description cs:.15,.5)},anchor=south},
    xlabel={\#MWTs in Training},
    ylabel={$\Delta$ Words \fone{}}]
     \addplot+[only marks, mark=*, draw=white, mark options={ fill=black},
     %nodes near coords={\name},%={\myvalue},
     % ... which requires to define a visualization dependency:
     %visualization depends on={\thisrow{name} \as \name},
     point meta=explicit symbolic,
     nodes near coords,
     every node near coord/.append style={color=black, font=\scriptsize},
           ]
           coordinates {
            % (0, 0) [Afrikaans-AfriBooms]
            % (0, 0) [Ancient\_Greek-Perseus]
            % (0, 0) [Ancient\_Greek-PROIEL]
            (12134, 4.19) [Arabic-PADT]
            % (0, 0) [Basque-BDT]
            % (0, 0) [Bulgarian-BTB]
            (641, 0.24) []%[Catalan-AnCora]
            % (0, 0) [Chinese-GSD]
            % (0, 0) [Croatian-SET]
            (18, 0) []%[Czech-CAC]
            (23, 0) []%[Czech-FicTree]
            (18, 0) [Czech-PDT]
            % (0, 0) [Danish-DDT]
            % (0, 0) [Dutch-Alpino]
            % (0, 0) [Dutch-LassySmall]
            % (0, 0) [English-EWT]
            % (0, 0) [English-GUM]
            % (0, 0) [English-LinES]
            % (0, 0) [Estonian-EDT]
            (38, 0) []%[Finnish-FTB]
            % (0, 0) [Finnish-TDT]
            (13, 0) []%[French-GSD]
            (13, 0) []%[French-Sequoia]
            % (0, 0) [French-Spoken]
            (844, 0.87) []%[Galician-CTG]
            (17, 0) []%[German-GSD]
            % (0, 0) [Gothic-PROIEL]
            (14, -0.01) []%[Greek-GDT]
            (14341, 10.08) [Hebrew-HTB]
            % (0, 0) [Hindi-HDTB]
            % (0, 0) [Hungarian-Szeged]
            % (0, 0) [Indonesian-GSD]
            (807, 0.27) []%[Italian-ISDT]
            (447, 0.34) [Italian-PoSTWITA]
            % (0, 0) [Japanese-GSD]
            % (0, 0) [Korean-GSD]
            % (0, 0) [Korean-Kaist]
            % (0, 0) [Latin-ITTB]
            % (0, 0) [Latin-PROIEL]
            % (0, 0) [Latvian-LVTB]
            % (0, 0) [Norwegian-Bokmaal]
            % (0, 0) [Norwegian-Nynorsk]
            % (0, 0) [Old\_Church\_Slavonic-PROIEL]
            % (0, 0) [Old\_French-SRCMF]
            (630, 0.15) []%[Persian-Seraji]
            % (0, 0) [Polish-LFG]
            (446, 0.76) []%[Polish-SZ]
            (767, 0.18) []%[Portuguese-Bosque]
            % (0, 0) [Romanian-RRT]
            % (0, 0) [Russian-SynTagRus]
            % (0, 0) [Serbian-SET]
            % (0, 0) [Slovak-SNK]
            % (0, 0) [Slovenian-SSJ]
            (933, 0.22) []%[Spanish-AnCora]
            % (0, 0) [Swedish-LinES]
            % (0, 0) [Swedish-Talbanken]
            (817, 1.23) [Turkish-IMST]
            % (0, 0) [Ukrainian-IU]
            % (0, 0) [Urdu-UDTB]
            % (0, 0) [Uyghur-UDT]
            % (0, 0) [Vietnamese-VTB]
           };
  \end{semilogxaxis}
\end{tikzpicture}
    \caption{Effect of the seq2seq component for MWT expansion in the tokenizer.}\label{fig:tok_seq2seq}
\end{figure}

\paragraph{Tokenizer.}
We perform ablation studies on the less standard components in the tokenizer, namely the gating mechanism in Eq.~(\ref{eqn:tokenizer_gating}) (\textit{gating}), the convolutional residual connections (\textit{conv}), and the seq2seq model in the MWT expander (\textit{seq2seq}), on all 61 big treebanks. As can be seen in Table \ref{tab:tok_ablation}, all but the gating mechanism make noticeable differences in macro \fone{}. When taking a closer look, we find that both \textit{gating} and \textit{conv} show a mixed contribution to each treebank, and we could have improved overall performance further through treebank-level component selection.
% \footnote{Both showed signs of overfitting of hyperparameters: on the 12 treebanks we used to tune the tokenizer, the gating component showed gains of 0.01, 0.07, and 0.04 \fone{} in Tokens, Sentences, and Words evaluation, and the convolutional component showed gains of 0.07, 0.32, and 2.46 \fone{}, respectively.}
%that although \textit{gating} improved the system's performance on some treebanks (\emph{e.g.}, Italian-PoSTWITA, $+2.24$ Sents \fone{}), it also negatively impacted others (\emph{e.g.}, Gothic-PROIEL, $-1.84$ Sents \fone{}).
One surprising discovery is that \textit{conv} greatly helps identify MWTs in Hebrew ($+34.89$ Words \fone{}) and sentence breaks in Ancient Greek-PROIEL ($+18.77$ Sents \fone{}).
%For these two components, we used one global configuration for all treebanks, which could have resulted in suboptimal performance for some. We note that an improvement through treebank-level hyperparameter tuning is possible.
In the case of \textit{seq2seq}, although the overall macro difference is small, it helps with the word segmentation performance on all treebanks where it makes any meaningful difference, most notably $+10.08$ on Hebrew and $+4.19$ on Arabic in Words \fone{} (see also Figure \ref{fig:tok_seq2seq}).
Finally, we note that \emph{dropout} plays an important role in safeguarding the tokenizer from overfitting.

\begin{figure*}[t!]
  \centering
  \setlength{\abovecaptionskip}{-6pt}
	\pgfplotstableread[row sep=\\,col sep=&]{
  treebank & short & total & seq2seq & identity & lowercase \\
  Ancient\_Greek-PROIEL & grc\_proiel & 13652 & 0.8378 & 0.1616 & 0.0006 \\
  Arabic-PADT & ar\_padt & 30239 & 0.8305 & 0.1695 & 0.0000 \\
  Ancient\_Greek-Perseus & grc\_perseus & 22135 & 0.7750 & 0.2250 & 0.0000 \\
  Korean-Kaist & ko\_kaist & 25278 & 0.7214 & 0.2786 & 0.0000 \\
  % Old\_Church\_Slavonic-PROIEL & cu\_proiel & 10100 & 0.7017 & 0.2967 & 0.0016 \\
  Latin-PROIEL & la\_proiel & 13939 & 0.6119 & 0.3799 & 0.0082 \\
  % Korean-GSD & ko\_gsd & 11958 & 0.6087 & 0.3913 & 0.0000 \\
  % Gothic-PROIEL & got\_proiel & 10114 & 0.5792 & 0.4197 & 0.0011 \\
  % Latin-Perseus & la\_perseus & 2276 & 0.5685 & 0.4192 & 0.0123 \\
  % Greek-GDT & el\_gdt & 10443 & 0.5511 & 0.4183 & 0.0306 \\
  Finnish-TDT & fi\_tdt & 18311 & 0.5464 & 0.4279 & 0.0257 \\
  % Latin-ITTB & la\_ittb & 10331 & 0.5336 & 0.4664 & 0.0000 \\
  Finnish-FTB & fi\_ftb & 15724 & 0.5194 & 0.4257 & 0.0549 \\
  % Serbian-SET & sr\_set & 10099 & 0.4958 & 0.4751 & 0.0291 \\
  Croatian-SET & hr\_set & 19543 & 0.4955 & 0.4781 & 0.0265 \\
  % North\_Sami-Giella & sme\_giella & 2161 & 0.4947 & 0.4526 & 0.0528 \\
  % Uyghur-UDT & ug\_udt & 10644 & 0.4939 & 0.5061 & 0.0000 \\
  Estonian-EDT & et\_edt & 37219 & 0.4922 & 0.4781 & 0.0298 \\
  Slovenian-SSJ & sl\_ssj & 14063 & 0.4919 & 0.4719 & 0.0361 \\
  Basque-BDT & eu\_bdt & 24095 & 0.4713 & 0.4896 & 0.0391 \\
  % Polish-SZ & pl\_sz & 10262 & 0.4667 & 0.4734 & 0.0599 \\
  % Czech-CAC & cs\_cac & 10912 & 0.4666 & 0.4994 & 0.0341 \\
  Slovak-SNK & sk\_snk & 12440 & 0.4566 & 0.4482 & 0.0952 \\
  Latvian-LVTB & lv\_lvtb & 14637 & 0.4445 & 0.5100 & 0.0455 \\
  % Turkish-IMST & tr\_imst & 10011 & 0.4385 & 0.4993 & 0.0622 \\
  Czech-PDT & cs\_pdt & 159284 & 0.4289 & 0.5323 & 0.0388 \\
  Romanian-RRT & ro\_rrt & 17074 & 0.4264 & 0.5517 & 0.0219 \\
  Polish-LFG & pl\_lfg & 13105 & 0.4194 & 0.5095 & 0.0711 \\
  Russian-SynTagRus & ru\_syntagrus & 118691 & 0.4175 & 0.5185 & 0.0639 \\
  Czech-FicTree & cs\_fictree & 16710 & 0.4159 & 0.5403 & 0.0438 \\
  Norwegian-Nynorsk & no\_nynorsk & 31250 & 0.4110 & 0.5470 & 0.0420 \\
  Norwegian-Bokmaal & no\_bokmaal & 36369 & 0.4077 & 0.5464 & 0.0459 \\
  % Slovenian-SST & sl\_sst & 2611 & 0.4075 & 0.5925 & 0.0000 \\
  % Ukrainian-IU & uk\_iu & 10371 & 0.4041 & 0.5494 & 0.0465 \\
  Bulgarian-BTB & bg\_btb & 16089 & 0.3901 & 0.5266 & 0.0833 \\
  % Swedish-Talbanken & sv\_talbanken & 9799 & 0.3823 & 0.5890 & 0.0287 \\
  % Russian-Taiga & ru\_taiga & 1253 & 0.3655 & 0.5515 & 0.0830 \\
  French-Spoken & fr\_spoken & 10010 & 0.3383 & 0.6617 & 0.0000 \\
  % Norwegian-NynorskLIA & no\_nynorsklia & 393 & 0.3282 & 0.6718 & 0.0000 \\
  % French-Sequoia & fr\_sequoia & 10011 & 0.3253 & 0.6485 & 0.0262 \\
  Italian-ISDT & it\_isdt & 11908 & 0.3236 & 0.6469 & 0.0295 \\
  Galician-CTG & gl\_ctg & 29772 & 0.3210 & 0.6357 & 0.0433 \\
  German-GSD & de\_gsd & 12486 & 0.3209 & 0.6438 & 0.0353 \\
  % Galician-TreeGal & gl\_treegal & 2025 & 0.3160 & 0.6237 & 0.0602 \\
  Catalan-AnCora & ca\_ancora & 56482 & 0.3118 & 0.6710 & 0.0172 \\
  Swedish-LinES & sv\_lines & 16462 & 0.3054 & 0.6474 & 0.0472 \\
  % Hungarian-Szeged & hu\_szeged & 11418 & 0.3017 & 0.6682 & 0.0300 \\
  % Danish-DDT & da\_ddt & 10332 & 0.3009 & 0.6576 & 0.0415 \\
  Hindi-HDTB & hi\_hdtb & 35217 & 0.2973 & 0.7027 & 0.0000 \\
  % Irish-IDT & ga\_idt & 1542 & 0.2964 & 0.6582 & 0.0454 \\
  French-GSD & fr\_gsd & 35768 & 0.2958 & 0.6756 & 0.0286 \\
  Spanish-AnCora & es\_ancora & 52336 & 0.2914 & 0.6892 & 0.0194 \\
  % Hebrew-HTB & he\_htb & 11408 & 0.2836 & 0.7164 & 0.0000 \\
  % Portuguese-Bosque & pt\_bosque & 10851 & 0.2759 & 0.6917 & 0.0323 \\
  Urdu-UDTB & ur\_udtb & 14581 & 0.2716 & 0.7284 & 0.0000 \\
  Italian-PoSTWITA & it\_postwita & 12335 & 0.2534 & 0.6809 & 0.0657 \\
  English-LinES & en\_lines & 17102 & 0.2088 & 0.7114 & 0.0798 \\
  Persian-Seraji & fa\_seraji & 15832 & 0.1902 & 0.8098 & 0.0000 \\
  % Dutch-Alpino & nl\_alpino & 11541 & 0.1894 & 0.7578 & 0.0528 \\
  % Dutch-LassySmall & nl\_lassysmall & 11397 & 0.1830 & 0.7618 & 0.0552 \\
  English-GUM & en\_gum & 13164 & 0.1636 & 0.7947 & 0.0418 \\
  English-EWT & en\_ewt & 25150 & 0.1488 & 0.7798 & 0.0715 \\
  % Afrikaans-AfriBooms & af\_afribooms & 5317 & 0.1307 & 0.8273 & 0.0419 \\
  Japanese-GSD & ja\_gsd & 11556 & 0.1211 & 0.8789 & 0.0000 \\
  Indonesian-GSD & id\_gsd & 12612 & 0.0696 & 0.7144 & 0.2160 \\
  Chinese-GSD & zh\_gsd & 12663 & 0.0018 & 0.9982 & 0.0000 \\
}\editratio

\begin{tikzpicture}[font=\scriptsize]
  \begin{axis}[
  ybar stacked,
  height=1.4in,
  bar width=5pt,
  width=0.88\textwidth,
  xtick=data,
  x tick label style={rotate=40, anchor=east, font=\tiny},
  tick align=inside,
  ylabel={Ratio of edit types},
  y label style={at={(axis description cs:.05,.5)},anchor=south},
  xticklabels from table={\editratio}{treebank},
  enlargelimits=0.03,
  % legend style={at={(0.5,-0.7)},
    % anchor=north,legend columns=-1},
  legend pos = outer north east,
  ]
\addplot table[x expr=\coordindex, y=seq2seq]{\editratio};
\addplot table[x expr=\coordindex, y=identity]{\editratio};
\addplot table[x expr=\coordindex, y=lowercase]{\editratio};
\legend{seq2seq, identity, lowercase};
  \end{axis}
  \end{tikzpicture}
  \vspace{-1.5em}
  \caption{Edit operation types as output by the \emph{edit} classifier on the official dev set. Due to space limit only treebanks containing over 120k dev words are shown and sorted by the ratio of \emph{seq2seq} operation.}
  \label{fig:edit-ratio}
\end{figure*}
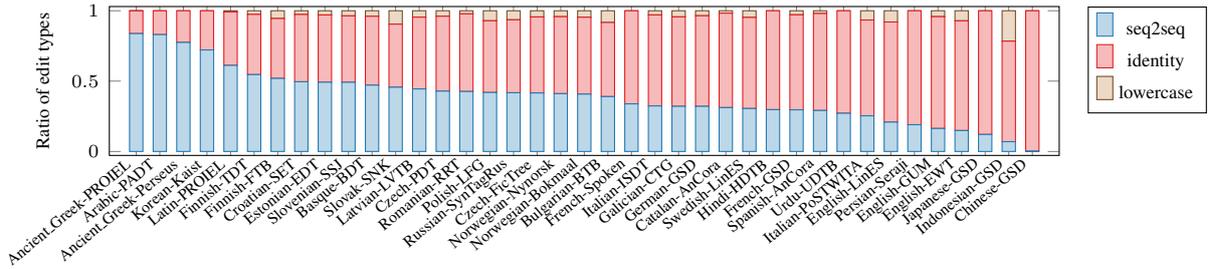

\begin{table}[t]
  \centering
  \small
  \setlength{\tabcolsep}{0.4em}
  \begin{tabular}{lccccc}
  \toprule
  System & UPOS & XPOS & UFeats & AllTags & PMI\\
  \midrule
  Stanford & \textbf{96.50} & \textbf{95.87}$^{\phantom{*}}$ & \textbf{95.01}$^{\phantom{***}}$ & \textbf{92.52}$^{\phantom{***}}$ & {\small \textbf{.0514}$\phantom{^*}$}\\
  $-$ \textit{biaff} & 96.47 & 95.71$^{*}$ & 94.13$^{***}$ & 91.32$^{***}$ & {\small .0497$^{*}$}\\
  \bottomrule
  \end{tabular}
  \caption{Ablation results for the tagger. All metrics are macro-averaged dev \fone{}, except PMI, which is explained in detail in the main text.}%the macro-averaged pointwise mutual information between AllTags and the other metrics.}
  \label{tab:tagger_ablation}
  \end{table}

\paragraph{POS/UFeats Tagger.}
The main novelty in our tagger is the explicit conditioning of XPOS and UFeats predictions on the UPOS prediction. % through a \emph{biaffine} classifier.
We compare this against a tagger that simply shares the hidden features between the UPOS, XPOS, and UFeats classifiers.
Since we used full-rank tensors in the biaffine classifier, treebanks with large, composite XPOS tagsets would incur prohibitive memory requirements.
We therefore exclude treebanks that either have more than 250 XPOS tags or don't use them, leaving 36 treebanks for this analysis.
We also measure consistency between tags by their pointwise mutual information
\begin{align*}
\text{PMI}=& \log\left(\frac{p_c(\text{AllTags})}{p_c(\text{UPOS})p_c(\text{XPOS})p_c(\text{UFeats})}\right),
\end{align*}
where $p_c(X)$ is the accuracy of $X$.
This quantifies (in nats) how much more likely it is to get all tags right than we would expect given their individual accuracies, if they were independent.
As can be seen in Table \ref{tab:tagger_ablation}, the added parameters do not affect UPOS performance significantly, but do help improve XPOS and UFeats prediction. Moreover, the biaffine classifier is markedly more consistent than the affine one with shared representations.

\begin{table}
\centering
\small
\setlength{\tabcolsep}{0.3em}
\begin{tabular}{lcccc}
\toprule
System & Big & Small & LowRes & All \\
\midrule
Stanford & \textbf{96.56}$^{\phantom{***}}$ & 91.72$^{*\phantom{**}}$ & \textbf{69.21}$^{\phantom{***}}$ & \textbf{94.22}$^{\phantom{***}}$ \\
$-$ \textit{edit \& seq2seq} & 89.97$^{{***}}$ & 82.68$^{{***}}$ & 63.50$^{**\phantom{*}}$ & 87.45$^{{***}}$\\
$-$ \textit{edit} & 96.48$^{*\phantom{**}}$ & \textbf{91.80}$^{\phantom{***}}$ & 68.30$^{\phantom{***}}$ & 94.10$^{\phantom{***}}$\\
$-$ \textit{dictionaries} & 95.37$^{{***}}$ & 90.43$^{{***}}$ & 66.02$^{*\phantom{**}}$ & 92.89$^{{***}}$\\
\bottomrule
\end{tabular}
\caption{Ablation results for the lemmatizer, split by different groups of treebanks. All metrics in the table are macro-averaged dev \fone{}.}
\label{tab:lemma_ablation}
\end{table}

\paragraph{Lemmatizer.}
We perform ablation studies on three individual components in our lemmatizer: the edit classifier (\textit{edit}), the sequence-to-sequence module (\textit{seq2seq}) and the dictionaries (\textit{dictionaries}).
As shown in Table~\ref{tab:lemma_ablation}, we find that our lemmatizer with all components achieves the best overall performance.
Specifically, adding the neural components (i.e., \emph{edit \& seq2seq}) drastically improves overall lemmatization performance over a simple dictionary-based approach ($+6.77$ \fone), and the gains are consistent over different treebank groups.
While adding the edit classifier slightly decreases the \fone{} score on small treebanks, it improves the performance on low-resource languages substantially ($+0.91$ \fone), and therefore leads to an overall gain of 0.11 \fone.
Treebanks where the largest gains are observed include Upper\_Sorbian-UFAL ($+4.55$ \fone), Kurmanji-MG ($+2.27$ \fone) and English-LinES ($+2.16$ \fone).
Finally, combining the neural lemmatizer with dictionaries helps capture common lemmatization patterns seen during training, leading to substantial improvements on all treebank groups.

To further understand the behavior of the edit classifier, for each treebank we present the ratio of all predicted edit types on dev set words in Figure~\ref{fig:edit-ratio}. We find that the behavior of the edit classifier aligns well with linguistic knowledge. For example, while Ancient Greek, Arabic and Korean require a lot of complex edits in lemmatization, the vast majority of operations in Chinese and Japanese are simple identity mappings.

% order: -edit&seq2seq -edit Stanford -dict
% p-values big: [9.99990000099999e-06, 0.04919950800491995, 9.99990000099999e-06]
% p-values small: [9.99990000099999e-06, 0.4010459895401046, 9.99990000099999e-06]
% p-values lowres: [0.0033699663003369964, 0.2009579904200958, 0.02209977900220998]
% p-values all: [9.99990000099999e-06, 0.06481935180648193, 9.99990000099999e-06]

\begin{table}
\centering
\small
\begin{tabular}{lcc}
\toprule
System & LAS & CLAS\\
\midrule
Stanford & \textbf{87.60}$^{\phantom{***}}$ & \textbf{84.68}$^{\phantom{***}}$\\
$-$ \textit{linearization} & 87.55$^{*\phantom{**}}$ & 84.62$^{*\phantom{**}}$\\
$-$ \textit{distance} & 87.43$^{{***}}$ & 84.48$^{{***}}$\\
%$-$ \textit{both} & 87.34$^{{***}}$ & 84.40$^{{***}}$\\
\bottomrule
\end{tabular}
\caption{Ablation results for the parser. All metrics in the table are macro-averaged dev \fone{}.}
\label{tab:parser_ablation}
\end{table}

\paragraph{Dependency Parser.}
The main innovation for the parsing module is terms that model locations of a dependent word relative to possible head words in the sentence. Here we examine the impact of these terms, namely linearization (Eq.~(\ref{eqn:linearization})) and distance (Eq.~(\ref{eqn:distance})). For this analysis, we exclude six treebanks with very small dev sets. As can be seen in Table \ref{tab:parser_ablation}, both terms contribute significantly to the final parser performance, with the distance term contributing slightly more.

\begin{figure}
  \centering
  \pgfplotstableread[row sep=\\,col sep=&]{
	model&	UPOS&	XPOS&	UFeats&	AllTags&	Lemmas&	UAS&	LAS&	CLAS&	MLAS&	BLEX\\
unofficial&	89.95&	86.50&	86.20&	80.36&	89.17&	79.04&	74.16&	70.39&	62.08&	65.28\\
 goldtok&	91.23246914&	87.35135802&	88.44506173&	80.9891358&	91.48222222&	81.89160494&	76.35296296&	72.62518519&	63.91975309&	67.2245679\\
 goldtag&	100&	100&	100&	100&	92.35890244&	85.12073171&	80.54890244&	77.20658537&	75.35378049&	70.90158537\\
 goldlemma&	100&	100&	100&	100&	100&	85.20682927&	80.67487805&	77.28536585&	75.43780488&	77.28536585\\
 goldparse&	100&	100&	100&	100&	100&	100&	100&	100&	100&	100\\
}\pipeline

\begin{tikzpicture}[font=\small, every mark/.append style={solid}]
\begin{axis}[
symbolic x coords={unofficial, goldtok, goldtag, goldlemma, goldparse},
enlargelimits=0.05,
xticklabels={0, Stanford+, +gold tok, +gold tag, +gold lemma, +gold parse},
x tick label style={rotate=15, anchor=north},
x label style={at={(axis description cs:0.5,-0.0)},anchor=north},
y label style={at={(axis description cs:.15,.5)},anchor=south},
xlabel={System},
ylabel={Test \fone},
ymin=60.5,
ymax=100,
height=5.14cm,
width=6cm,
legend pos=south east,
legend style={at={(1,1)},anchor=north west, font=\tiny},
legend columns=1,
ymajorgrids=true,
grid style=dashed,
]
\addplot[densely dashed, color=orange, mark=star] table[x=model,y=UPOS]{\pipeline};
\addplot[densely dashed, color=purple, mark=o] table[x=model,y=XPOS]{\pipeline};
\addplot[densely dashed, color=blue, mark=x] table[x=model,y=UFeats]{\pipeline};
\addplot[densely dashed, color=green, mark=+] table[x=model,y=AllTags]{\pipeline};
\addplot[dotted, color=red, mark=star] table[x=model,y=Lemmas]{\pipeline};
\addplot[color=orange, mark=o] table[x=model,y=UAS]{\pipeline};
\addplot[color=purple, mark=x] table[x=model,y=CLAS]{\pipeline};
\addplot[thick, color=blue, mark=+] table[x=model,y=LAS]{\pipeline};
\addplot[thick, color=green, mark=star] table[x=model,y=MLAS]{\pipeline};
\addplot[thick, color=red, mark=o] table[x=model,y=BLEX]{\pipeline};
\legend{UPOS, XPOS, UFeats, AllTags, Lemmas, UAS, CLAS, LAS, MLAS, BLEX}
\end{axis}
\end{tikzpicture}
  \vspace{-1.2em}
  \caption{Pipeline ablation results. Dashed, dotted, and solid lines represent tagger, lemmatizer, and parser metrics, respectively. Official evaluation metrics are highlighted with thickened lines.} \label{fig:pipeline_ablation}
\end{figure}
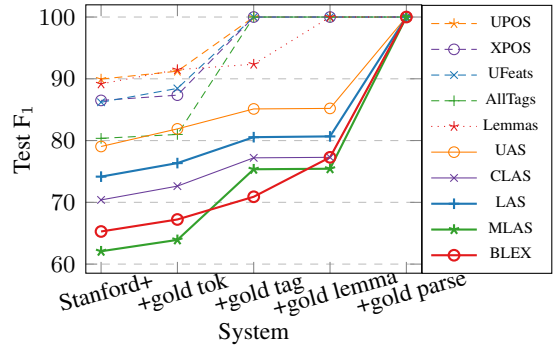

\paragraph{Pipeline Ablation.} We analyze the contribution of each pipeline component by incrementally replacing them with gold annotations and observing  performance change.
As shown in Figure \ref{fig:pipeline_ablation}, most downstream systems benefit moderately from gold sentence and word segmentation, while the parser largely only benefits from improved POS/UFeats tagger performance (aside from BLEX, which is directly related to lemmatization performance and benefits notably).
% Finally, we note that even given gold annotations from all upstream tasks, the parser still has plenty of room for improvement.
Finally, we note that the parser still is far from perfect even given gold annotations from all upstream tasks, but our components in the pipeline are very effective at closing the gap between predicted and gold annotations.

\section{Conclusion \& Future Directions}

In this paper, we presented Stanford's submission to the \udst. Our submission consists of neural components for each stage of a pipeline from raw text to dependency parses. %The components feature innovative approaches for %multi-resolution modeling of character sequences,
%robust combination of statistical symbolic knowledge and neural systems, consistency enhancements for POS tagging and morphological feature prediction, as well as arc distance and linear order modeling in dependency parsing.
%The resulting pipeline system achieves very competitive results on big treebanks.
The final system was very competitive on big treebanks; after fixing our preprocessing bug, it would have outperformed all official systems on all metrics for low-resource treebank categories.

One of the greatest opportunities for further gains is through the use of context-sensitive word embeddings, such as ELMo \cite{peters2018deep} and ULMfit \cite{howard2018universal}.
Although this requires a large resource investment, HIT-SCIR \cite{che18towards} has shown solid improvements from incorporating these embeddings.

\bibliography{udst2018,K17-3}

\begin{thebibliography}{}
\expandafter\ifx\csname natexlab\endcsname\relax\def\natexlab#1{#1}\fi

\bibitem[{Bahdanau et~al.(2015)Bahdanau, Cho, and Bengio}]{bahdanau2014neural}
Dzmitry Bahdanau, Kyunghyun Cho, and Yoshua Bengio. 2015.
\newblock Neural machine translation by jointly learning to align and
  translate.
\newblock {\em ICLR\/} .

\bibitem[{Bojanowski et~al.(2017)Bojanowski, Grave, Joulin, and
  Mikolov}]{bojanowski2016enriching}
Piotr Bojanowski, Edouard Grave, Armand Joulin, and Tomas Mikolov. 2017.
\newblock \href{http://aclweb.org/anthology/Q17-1010}{Enriching word vectors
  with subword information}.
\newblock {\em Transactions of the Association for Computational Linguistics\/}
  5:135--146.
\newblock
  \href{http://aclweb.org/anthology/Q17-1010}{http://aclweb.org/anthology/Q17-1010}.

\bibitem[{Che et~al.(2018)Che, Liu, Wang, and Liu}]{che18towards}
Wanxiang Che, Yijia Liu, Yuxuan~Zheng Bo~Wang, and Ting Liu. 2018.
\newblock Towards better {UD} parsing: Deep contextualized word embeddings,
  ensemble, and treebank concatenation.
\newblock In {\em {Proceedings of the CoNLL 2018 Shared Task: Multilingual
  Parsing from Raw Text to Universal Dependencies}\/}.

\bibitem[{Chen et~al.(2017)Chen, Huang, Chiang, and Chen}]{chen2017improved}
Huadong Chen, Shujian Huang, David Chiang, and Jiajun Chen. 2017.
\newblock Improved neural machine translation with a syntax-aware encoder and
  decoder.
\newblock In {\em Proceedings of the 55th Annual Meeting of the Association for
  Computational Linguistics\/}.

\bibitem[{Chu and Liu(1965)}]{chu1965shortest}
Yoeng-Jin Chu and Tseng-Hong Liu. 1965.
\newblock On the shortest arborescence of a directed graph.
\newblock {\em Scientia Sinica\/} 14:1396--1400.

\bibitem[{Dozat et~al.(2017)Dozat, Qi, and
  Manning}]{dozat-qi-manning:2017:K17-3}
Timothy Dozat, Peng Qi, and Christopher~D. Manning. 2017.
\newblock \href{http://www.aclweb.org/anthology/K/K17/K17-3002.pdf}{Stanford's
  graph-based neural dependency parser at the {CoNLL 2017 Shared Task}}.
\newblock In {\em Proceedings of the CoNLL 2017 Shared Task: Multilingual
  Parsing from Raw Text to Universal Dependencies\/}. pages 20--30.
\newblock
  \href{http://www.aclweb.org/anthology/K/K17/K17-3002.pdf}{http://www.aclweb.org/anthology/K/K17/K17-3002.pdf}.

\bibitem[{Edmonds(1967)}]{edmonds1967optimum}
Jack Edmonds. 1967.
\newblock Optimum branchings.
\newblock {\em Journal of Research of the National Bureau of Standards\/}
  71:233--240.

\bibitem[{Gal and Ghahramani(2016)}]{gal2016dropout}
Yarin Gal and Zoubin Ghahramani. 2016.
\newblock Dropout as a {Bayesian} approximation: Representing model uncertainty
  in deep learning.
\newblock In {\em International Conference on Machine Learning\/}. pages
  1050--1059.

\bibitem[{He et~al.(2016)He, Zhang, Ren, and Sun}]{he2016residual}
Kaiming He, Xiangyu Zhang, Shaoqing Ren, and Jian Sun. 2016.
\newblock Deep residual learning for image recognition.
\newblock In {\em {CVPR}\/}.

\bibitem[{Howard and Ruder(2018)}]{howard2018universal}
Jeremy Howard and Sebastian Ruder. 2018.
\newblock Universal language model fine-tuning for text classification.
\newblock In {\em Proceedings of the 56th Annual Meeting of the Association for
  Computational Linguistics\/}.

\bibitem[{Kingma and Ba(2015)}]{kingma2015adam}
Diederik~P Kingma and Jimmy Ba. 2015.
\newblock {A}dam: A method for stochastic optimization.
\newblock {\em ICLR\/} .

\bibitem[{Marcheggiani and Titov(2017)}]{marcheggiani2017encoding}
Diego Marcheggiani and Ivan Titov. 2017.
\newblock Encoding sentences with graph convolutional networks for semantic
  role labeling.
\newblock In {\em Proceedings of the Conference on Empirical Methods for
  Natural Language Processing\/}.

\bibitem[{Mikolov et~al.(2013)Mikolov, Sutskever, Chen, Corrado, and
  Dean}]{mikolov2013distributed}
Tomas Mikolov, Ilya Sutskever, Kai Chen, Greg~S Corrado, and Jeff Dean. 2013.
\newblock Distributed representations of words and phrases and their
  compositionality.
\newblock In {\em Advances in Neural Information Processing Systems\/}. pages
  3111--3119.

\bibitem[{Peters et~al.(2018)Peters, Neumann, Iyyer, Gardner, Clark, Lee, and
  Zettlemoyer}]{peters2018deep}
Matthew~E Peters, Mark Neumann, Mohit Iyyer, Matt Gardner, Christopher Clark,
  Kenton Lee, and Luke Zettlemoyer. 2018.
\newblock Deep contextualized word representations.

\bibitem[{Reddi et~al.(2018)Reddi, Kale, and Kumar}]{reddi2018convergence}
Sashank~J Reddi, Satyen Kale, and Sanjiv Kumar. 2018.
\newblock On the convergence of {A}dam and beyond.
\newblock {\em ICLR\/} .

\bibitem[{Serban et~al.(2017)Serban, Klinger, Tesauro, Talamadupula, Zhou,
  Bengio, and Courville}]{serban2017multiresolution}
Iulian~Vlad Serban, Tim Klinger, Gerald Tesauro, Kartik Talamadupula, Bowen
  Zhou, Yoshua Bengio, and Aaron~C Courville. 2017.
\newblock Multiresolution recurrent neural networks: An application to dialogue
  response generation.
\newblock In {\em AAAI\/}. pages 3288--3294.

\bibitem[{Srivastava et~al.(2015)Srivastava, Greff, and
  Schmidhuber}]{srivastava2015highway}
Rupesh~Kumar Srivastava, Klaus Greff, and J{\"u}rgen Schmidhuber. 2015.
\newblock Highway networks.
\newblock In {\em Proceedings of the Deep Learning Workshop at the
  International Conference on Machine Learning\/}.

\bibitem[{Zeman et~al.(2018)Zeman, Haji{\v{c}}, Popel, Potthast, Straka,
  Ginter, Nivre, and Petrov}]{udst:overview}
Daniel Zeman, Jan Haji{\v{c}}, Martin Popel, Martin Potthast, Milan Straka,
  Filip Ginter, Joakim Nivre, and Slav Petrov. 2018.
\newblock {CoNLL 2018 Shared Task: Multilingual Parsing from Raw Text to
  Universal Dependencies}.
\newblock In {\em Proceedings of the CoNLL 2018 Shared Task: Multilingual
  Parsing from Raw Text to Universal Dependencies\/}. Association for
  Computational Linguistics, Brussels, Belgium, pages 1--20.

\bibitem[{Zeman et~al.(2017)Zeman, Popel, Straka, Hajic, Nivre, Ginter,
  Luotolahti, Pyysalo, Petrov, Potthast, Tyers, Badmaeva, Gokirmak, Nedoluzhko,
  Cinkova, Hajic~jr., Hlavacova, Kettnerov\'{a}, Uresova, Kanerva, Ojala,
  Missil\"{a}, Manning, Schuster, Reddy, Taji, Habash, Leung, de~Marneffe,
  Sanguinetti, Simi, Kanayama, dePaiva, Droganova, Mart\'{i}nez~Alonso,
  \c{C}\"{o}ltekin, Sulubacak, Uszkoreit, Macketanz, Burchardt, Harris,
  Marheinecke, Rehm, Kayadelen, Attia, Elkahky, Yu, Pitler, Lertpradit, Mandl,
  Kirchner, Alcalde, Strnadov\'{a}, Banerjee, Manurung, Stella, Shimada, Kwak,
  Mendonca, Lando, Nitisaroj, and Li}]{udst:overview2017}
Daniel Zeman, Martin Popel, Milan Straka, Jan Hajic, Joakim Nivre, Filip
  Ginter, Juhani Luotolahti, Sampo Pyysalo, Slav Petrov, Martin Potthast,
  Francis Tyers, Elena Badmaeva, Memduh Gokirmak, Anna Nedoluzhko, Silvie
  Cinkova, Jan Hajic~jr., Jaroslava Hlavacova, V\'{a}clava Kettnerov\'{a},
  Zdenka Uresova, Jenna Kanerva, Stina Ojala, Anna Missil\"{a}, Christopher~D.
  Manning, Sebastian Schuster, Siva Reddy, Dima Taji, Nizar Habash, Herman
  Leung, Marie-Catherine de~Marneffe, Manuela Sanguinetti, Maria Simi, Hiroshi
  Kanayama, Valeria dePaiva, Kira Droganova, H\'{e}ctor Mart\'{i}nez~Alonso,
  \c{C}a\u{g}rı \c{C}\"{o}ltekin, Umut Sulubacak, Hans Uszkoreit, Vivien
  Macketanz, Aljoscha Burchardt, Kim Harris, Katrin Marheinecke, Georg Rehm,
  Tolga Kayadelen, Mohammed Attia, Ali Elkahky, Zhuoran Yu, Emily Pitler, Saran
  Lertpradit, Michael Mandl, Jesse Kirchner, Hector~Fernandez Alcalde, Jana
  Strnadov\'{a}, Esha Banerjee, Ruli Manurung, Antonio Stella, Atsuko Shimada,
  Sookyoung Kwak, Gustavo Mendonca, Tatiana Lando, Rattima Nitisaroj, and Josie
  Li. 2017.
\newblock {CoNLL 2017 Shared Task: Multilingual Parsing from Raw Text to
  Universal Dependencies}.
\newblock In {\em Proceedings of the CoNLL 2017 Shared Task: Multilingual
  Parsing from Raw Text to Universal Dependencies\/}. Association for
  Computational Linguistics, pages 1--19.

\bibitem[{Zhang et~al.(2018)Zhang, Qi, and Manning}]{zhang2018graph}
Yuhao Zhang, Peng Qi, and Christopher~D. Manning. 2018.
\newblock Graph convolution over pruned dependency trees improves relation
  extraction.
\newblock In {\em Proceedings of the Conference on Empirical Methods for
  Natural Language Processing\/}.

\end{thebibliography}
\bibliographystyle{acl_natbib}

\end{document}